\documentclass[cameraready]{Interspeech}

\title{Using embeddings to predict spoken word duration and pitch in Mandarin monosyllabic words}

\author[affiliation={1}]{Xiaoyun}{Jin}

\author[affiliation={2}]{Mirjam}{Ernestus}

\author[affiliation={1}]{R.Harald}{Baayen}

\address{\mbox{}}
\address{
    $^1$ Quantitative Linguistik, University of Tuebingen,  Germany $^2$  Center for Language Studies, Radboud University, 6525 HT Nijmegen, The Netherlands
}
\email{xiaoyun.jin@uni-tuebingen.de,  	mirjam.ernestus@ru.nl, harald.baayen@uni-tuebingen.de}


\usepackage{comment}

\begin{document}
\begin{CJK}{UTF8}{bsmi} 

\maketitle

\begin{abstract}

Time-normalized f0 contours of Mandarin words in conversational speech have been shown to be predictable in part from their contextualized embeddings (CEs). The present study investigates whether CEs also predict spoken word duration for  7470 tokens of Mandarin monosyllabic CV words extracted from a Mandarin corpus of spontaneous speech. We show that CEs indeed are predictive for duration, above chance level, not only at the type level, but also at the level of individual tokens, as indicated by the results obtained with the type-wise and token-wise permutation baselines. We also show that the predicted durations are sufficiently precise to back-transform predicted f0 contours in [0,1] normalized time to contours on the ms time scale. The resulting predicted contours approximate empirical contours and also outperform a permutation baseline.  
\end{abstract}

\section{Introduction}

Prosody concerns those phonetic properties that are not covered by words' vowels and consonants, such as spoken word duration, f0 contour and prominence \cite{cole2015prosody}.  The realization of prosodic properties is governed by a wide range of factors, such as the prosodic properties of neighbouring words \cite{shen1990prosody}, internal and external sandhi processes \cite{shih1986prosodic,niebuhr2011segment}, frequency of use \cite{gahl2008time}, word category \cite{lohmann2018cut}, speech rate \cite{tseng2004speech}, morphosyntactic function \cite{plag2015homophony}, and the emotional state of the speaker \cite{frick1985communicating}. 

For spoken word duration, previous work \cite{gahl_time_2024} has shown that the spoken word duration of  English homophones, a typical prosodic feature, is predictable to a considerable extent from the meanings of these homophones, while controlling for a wide range of other prosodic variables as well as inherent segment duration.  This study operationalized word meaning with embeddings from distributional semantics. It quantified the amount of support a word's segment receives from its embedding using the Discriminative Lexicon Model (DLM) \cite{Heitmeier:Chuang:Baayen:2025}, and reported this support is a strong predictor of spoken word duration, independently of word frequency and probability measures.   

Given these results for English,  the present study investigates whether words' meanings are also predictive for spoken word duration in Mandarin Chinese. Whereas standard word embeddings was used \cite{gahl_time_2024} to represent the meaning of homophone types, in the present study, we take this line of research a step further by considering whether contextualized embeddings obtained with a large language model, as approximations of the meanings of word tokens in utterance context, also predict tokens' spoken word durations. 

Embeddings have been reported to be predictive not only for spoken word duration (English), but also for f0 contours in Mandarin \cite{Chuang:Bell:Tseng:Baayen:2024,lu2025realization} and English \cite{chuangwords}. These studies used the Generalized Additive Model (GAM) \cite{wood2017generalized} to decompose empirical corpus-extracted pitch contours into component contours tied to a wide range of linguistic  variables including speaker, speech rate, neighbouring tones, sentence position, and the presence of pauses. In these studies, strong f0 components emerged that were tied to word type, indicating that words can have their own tonal signatures in f0 contours. The full empirical contours also carry an array of additional component signatures that are linked to a wide range of other prosodic factors \cite{Chuang:Bell:Tseng:Baayen:2024,lu2025realization}. 

The above-mentioned studies used the DLM to show that these word-specific pitch signatures can be predicted from their tokens' contextualized embeddings. Given an $n \times k$ matrix $\bm{S}$ representing $k$-dimensional contextualized embeddings for $n$ tokens, and an $n \times m$ matrix $\bm{C}$ with words' time-normalized f0 contours, a linear mapping $\bm{W}$, estimated by solving $\bm{S}\bm{W} = \bm{C}$ is used to obtain the predicted pitch contours $\hat{\bm{C}}$. When the mapping $\bm{W}$ is applied to the centroid of the contextualized embeddings associated with the tokens of a given word type, the f0 signature specific to that type (as estimated by a GAM \cite{wood2017generalized}) is obtained.  At a higher level of aggregation, when the centroid of the contextualized embeddings of all word tokens of all types sharing the same Mandarin tone is given as input to the mapping $\bm{W}$, the result is a predicted tone pattern for that tone that is highly similar to the tone pattern reconstructed by a GAM. 

This line of research has focused on the shape of words' pitch contours, modelling the relation between meaning (approximated with contextualized embeddings)  and pitch in normalized time.  The duration of the pitch contours was taken into account as a covariate in the GAMs, as tone contours and word duration are not independent \cite{ho1976acoustic}.  In the present study, we therefore also investigate to what extent words' f0 contours can be predicted from tokens' contextualized embeddings not in normalized time, but in real time.  If the durations predicted from contextualized embeddings are of sufficient quality, then these can be used to back-transformed predicted f0 contours in normalized time to predicted f0 contours in real time.

\section{Method}

\subsection{Data}

\noindent
The corpus used in the current study is the Taiwan Mandarin spontaneous speech corpus \cite{fon2004preliminary}, which provides word-level transcriptions using traditional Chinese characters. We followed the transcriptions in the corpus, and distinguished between word types on the basis of the characters with which the words are transcribed. 

In Mandarin, single-character words always correspond to single spoken syllables. The maximal structure of a syllable is CGVN: an onset consonant (C), a pre-nuclear glide (G), a vocalic nucleus (V) and a coda consonant (N), which, according to the phonotactic constraints on Mandarin syllables, must be a nasal \cite{duanmu2007phonology}. In Mandarin, the vowel can be one of seven monophthongs as well as one of eight diphthongs \cite{yi1920lectures}.  The present study is restricted to words with one of the vowels /a, i, u, \textschwa/ without coda consonants and glides, to be able to control for segmental effects on duration \cite{wu2015duration}.

Our dataset comprised 53,139 tokens of 699 word types with /a/, /i/, /u/, and /\textschwa/, of which 6481 tokens had an empty onset. About 85\% of these 53,139 tokens belong to just 32 high-frequency words, such as  的 \textit{de0}, `genitive',  你 \textit{ni3}, `you', and  他 \textit{ta1}, `he'.  In order to prevent model predictions from being heavily biased by the highest frequency words, we randomly sampled 220 tokens (4 tokens * 55 speakers) across 55 speakers for words with a token frequency greater than 220. As a further measure to ensure that for statistical analysis there are sufficient tokens for a word type, we also excluded those words with a token frequency lower than 10, which left us with 8,187 tokens. 

We used the Montreal Forced Aligner \cite{mcauliffe2017montreal} to determine the boundary between the consonant and vowel in the CV words under study. Among the 8,187 tokens, 408 tokens were not assigned a segmental boundary, typically due to these having extremely short durations. Forced-aligned text grids were therefore  audited manually in Praat \cite{boersma19922022} to ensure that segment boundaries were correctly placed. Due to unclear vowel pronunciations, caused by vowel reduction or background noise, 304 tokens were excluded during manual verification, which left us with 7,476 tokens of 102 word types.

That is, for each token, we calculated a context-sensitive embedding as an approximation of its meaning in context. The contextualized embeddings that we used were obtained with GPT-2 (for technical details, see \cite{Chuang:Bell:Tseng:Baayen:2024} and CKIP \footnote{ckiplab/gpt2-base-chinese, which is available on https://github.com/ckiplab/ckip-transformers.}). GPT-2 was applied to the utterances in the Taiwan Mandarin spontaneous speech corpus \cite{fon2004preliminary}, resulting in a $7,476 \times 768$ matrix $\bm{S}$ with contextualized embeddings.

For each word token, we extracted the duration of the vowel (resulting in a $7,476 \times 1$ matrix $\bf{C}_v$ of vowel durations) as well as the duration of the word (resulting in a $7,476 \times 1$ matrix $\bf{C}_w$ of word durations).  We also calculated time-normalized pitch contours for each of the tokens, following the procedure documented in detail in \cite{jin2025}. For this analysis, 9 word types with extremely short duration or octave jumps in f0 were removed from the analysis. The resulting dataset contains 6,118 tokens representing 93 word types, a subset of the data points considered in the present study. For each token, 100 pitch measurements in normalized time were retained, resulting in a $6,118 \times 100$ matrix $\bf{C}_p$.

\subsection{Procedure}

We obtained predicted durations and pitch contours by solving for $\textbf{W}$ the equations $\bm{S}\bm{W}_v = \bm{C}_v$, $\bm{S}\bm{W}_w = \bm{C}_w$, and $\bm{S}\bm{W}_p = \bm{C}_p$, and calculating
\begin{eqnarray}
    \hat{\bm{C}_v} &=& \bm{S}\bm{W}_v  \\ \nonumber 
    \hat{\bm{C}_w} &=& \bm{S}\bm{W}_w  \\ \nonumber 
    \hat{\bm{C}_p} &=& \bm{S}\bm{W}_p.  \label{eq:Ws}
\end{eqnarray}
To evaluate prediction quality, we made use of 10-fold cross-validation. For each of 10 runs, the data was split into 90\% training data and 10\% test data, in such a way that the number of tokens of any word type was proportionally represented in both the training and the test data.  For a given response variable, this  procedure yields 10 quality values for the training data and 10 quality values for the held-out data.  For the vowel and word duration response variables, we assessed prediction quality with the Pearson correlation of observed and predicted values.  For the predicted pitch contours in normalized time, we evaluated quality with a nearest neighbour accuracy measure. For each token, we used the mean squared error to find the token with the most similar empirical pitch contour. The predicted pitch contour of that token was counted as correct if the most similar token pitch contour belonged to a token of the same word type.  Across all tokens in training data, c.q. test data, this procedure results in an accuracy measure (percentage correct).  

Vowel and word duration, and f0 contours, are co-determined by many factors other than word meanings. Factors such as local speech rate, the emotional state of a speaker at the time of speaking, and the presence of neighbouring pauses, are unlikely to be strongly represented in the contextualized embeddings obtained with a large language model that was trained on huge volumes of written text and that represents an average, speaker-independent, estimate of what a word token means given its local preceding context.  Part of speech (word category) is reflected in embedding spaces, but only weakly so, whereas word meaning shows strong clustering in 2-D t-SNE  \cite{maaten2008visualizing} maps, as shown by \cite{Chuang:Bell:Tseng:Baayen:2024}.  In the present study, we evaluated prediction quality against permutation baselines that obliterate the relation between contextualized embeddings and duration or f0 contour.  In this way, we can ascertain whether the  predictions derived from the empirical embeddings are more precise than those in which the relation between form and meaning has been randomized. In the general discussion, we return to the question of to what extent prosodic factors are present in the embeddings.   

We considered two separate permutation baselines, a lenient one and a stringent one. The lenient permutation baseline, henceforth `global baseline', was obtained by randomly permuting the word identifiers.  Embeddings are randomly re-associated with tokens and their durations and pitch contours.  If there is no information in embeddings about duration and f0, then breaking the relation between form and meaning should not make a difference. If there is some alignment of the semantic and the form spaces, then the empirical cross-validation results should be superior to the results of the global permutation baseline. 

If the global permutation baseline underperforms compared to the empirical results, the contextualized embeddings of tokens are better aligned with the durations and f0 contours of their own types than those of other types.  But we can ask whether token-specific durations and f0 contours are better aligned with their own token-specific embeddings than with the embeddings of other tokens of the same type.  To assess whether there is a signal at the level of individual tokens, we implemented a second permutation baseline,  henceforth the `type-wise permutation baseline'.  For this baseline, we did not permute across types, but only within types. For any given type, we randomly permuted the embeddings of the tokens belonging to that type. If this stringent permutation baseline underperforms compared to the empirical results, this indicates that there is some alignment of form and meaning even at the within-word token level.

\section{Results}

\subsection{Training data}

\noindent
Under 10-fold cross-validation, the mean correlation for vowel duration in training was $0.535$, 
the mean correlation for the global permutation baseline was 0.339 and mean correlation for the type-wise permutation baseline was 0.506, lower than the empirical mean correlation ($t_{(9)} = -18.473, p < 0.0001$).  

With respect to word duration, the empirical mean correlation was $0.557$. 
The global permutation baseline was 0.334.  The type-wise permutation baseline correlation for word duration was 0.527, lower than the empirical correlation  ($t_{(9)} = -23.493, p < 0.0001$). For the f0 contours, the empirical mean accuracy was $0.184$, 
the global permutation baseline accuracy was 0.021, lower than the empirical accuracy ($t_{(9)} = -121.09, p < 0.0001$). However, the mean accuracy for the type-wise permutation baseline was 0.195, higher than the empirical mean accuracy ($t_{(9)} = 5.587, p = 0.0003)$.
%
%

\subsection{Held-out data}

\noindent
For the held-out data, the mean correlation for vowel duration was $0.366$, 
the mean correlation for the global permutation baseline was -0.011 and the mean correlation for the type-wise permutation baseline was 0.309, lower than the empirical mean correlation ($t_{(9)} = -4.314, p < 0.0020$).  

For word duration, the empirical mean correlation was $0.399$. 
The global permutation baseline was again close to zero 0.000), and the type-wise permutation baseline correlation was 0.341, lower than the empirical correlation  ($t_{(9)} = -5.839, p =  0.0002$).

As to the f0 contours, the empirical mean accuracy was $0.170$, 
the global permutation baseline accuracy was $0.026$, 
lower than the empirical accuracy  ($t_{(9)} = -35.758, p <  0.0001$), but the mean correlation for the type-wise permutation baseline was 0.180, higher but not significantly higher than the empirical accuracy ($t_{(9)} = 2.180, p = 0.057$).

These results indicate that vowel  durations, word durations, and f0 contours of Mandarin monosyllabic words can be predicted from their contextualized embeddings above chance levels, replicating similar results reported by \cite{gahl_time_2024} for English heterographic homophone duration, and for Mandarin f0 contours \cite{Chuang:Bell:Tseng:Baayen:2024,lu2025realization}. What the present study adds to these earlier studies is evidence that for duration, but not for f0 contours, meaning and phonetic realization are entangled even at the level of individual tokens, as indicated by the results obtained with the type-wise permutation baseline.


%


\subsection{Predicting pitch contours in real time}

%
%
%
%

\noindent
We have seen that both the shape of a word's pitch contour (in normalized time) and its actual duration (in ms) can be predicted from words' contextualized embeddings above chance level.  We now consider how good embedding-driven predictions are for the F0 contours in real time (ms). For f0 contours, these predictions are expected to be less accurate, given that the type-based permutation baseline yielded similar prediction accuracies as the token level. In what follows, we therefore consider performance for type embedding centroids and the corresponding GAM-predicted f0 contours.  Since many factors co-determining prosody are not taken into account when basing predictions on embeddings only, results are unavoidably imprecise. We therefore compare the embedding-derived predictions with a permutation baseline that destroys the link between types and embeddings.  For this analysis, the smaller dataset (93 word types) is used, which excludes word types with extremely short duration or octave jumps in pitch measurements.

For a given type, we calculated the centroid of its contextualized embeddings, and then used the appropriate mapping (see equation~\ref{eq:Ws}) to obtain the corresponding predicted vowel duration and the pitch contour for the vowel part of the word.  By multiplying the f0 measurement time points in normalized time by the corresponding predicted duration, we obtain predicted pitch contours for 93 word types in real time.

To evaluate the quality of these predicted f0 contours in real time, we proceeded in two steps.  We first determined which word token has a contextualized embedding that is closest to the centroid of its word type. We then compared the predicted word-specific pitch contour with the empirical pitch contour of this token, using Dynamic Time Warping (DTW) \cite{sakoe2003dynamic} to evaluate the distance between the observed and predicted time series of f0 values, which are of unequal length.  This procedure resulted in a distance measure for each of the 93 word types. 

Empirical distances were compared with a permutation baseline in which type-wise centroids were calculated after first randomly permuting embeddings.  These random centroids were then used to generate predicted f0 contours in real time, and DTW was again used to calculate the distances between observed and predicted f0 contours. A paired-samples $t$-test for the 93 Euclidean distances showed that the average distance for the empirical predictions is shorter than that of the permutation-based predictions ($t(92) = -3.2529$, $p = 0.001598$).  The results of the more stringent type-wise permutation baseline support the same conclusion ($t(92) = -2.6826$, $p = 0.00866$).

\begin{figure}
    \centering
    \includegraphics[width=1\linewidth]{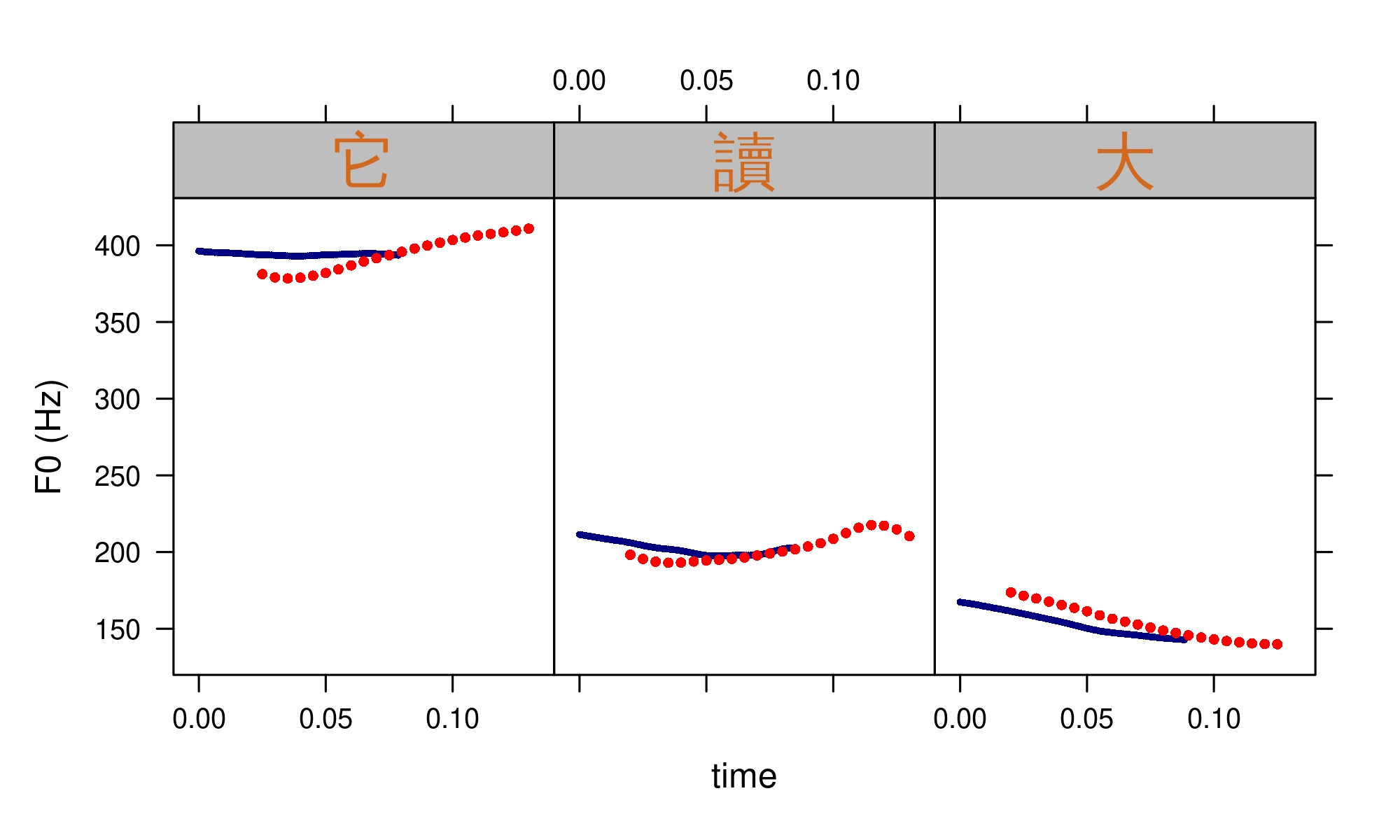}
    \caption{Examples of three Mandarin words that have the largest (它), medium (讀) and shortest (大) distance between observed (red) and predicted (blue) f0 contours. 
    }
    \label{fig:t3}
\end{figure}

Figure~\ref{fig:t3} presents the empirical and predicted pitch contours for the three words with the longest, the medium and the shortest Euclidean distance. For 它 (left panel), both shape and duration are off. For 讀 (middle panel), the shape is better approximated. For 大 (right panel), the shape is well-estimated, but the onset of the predicted f0 contour is too early. Taking the many non-semantic factors shaping prosody into account will make it possible to improve prediction accuracy  substantially, see, e.g., \cite{lu2025realization}. 

\begin{figure}[h]
    \centering
    \includegraphics[width=1\linewidth]{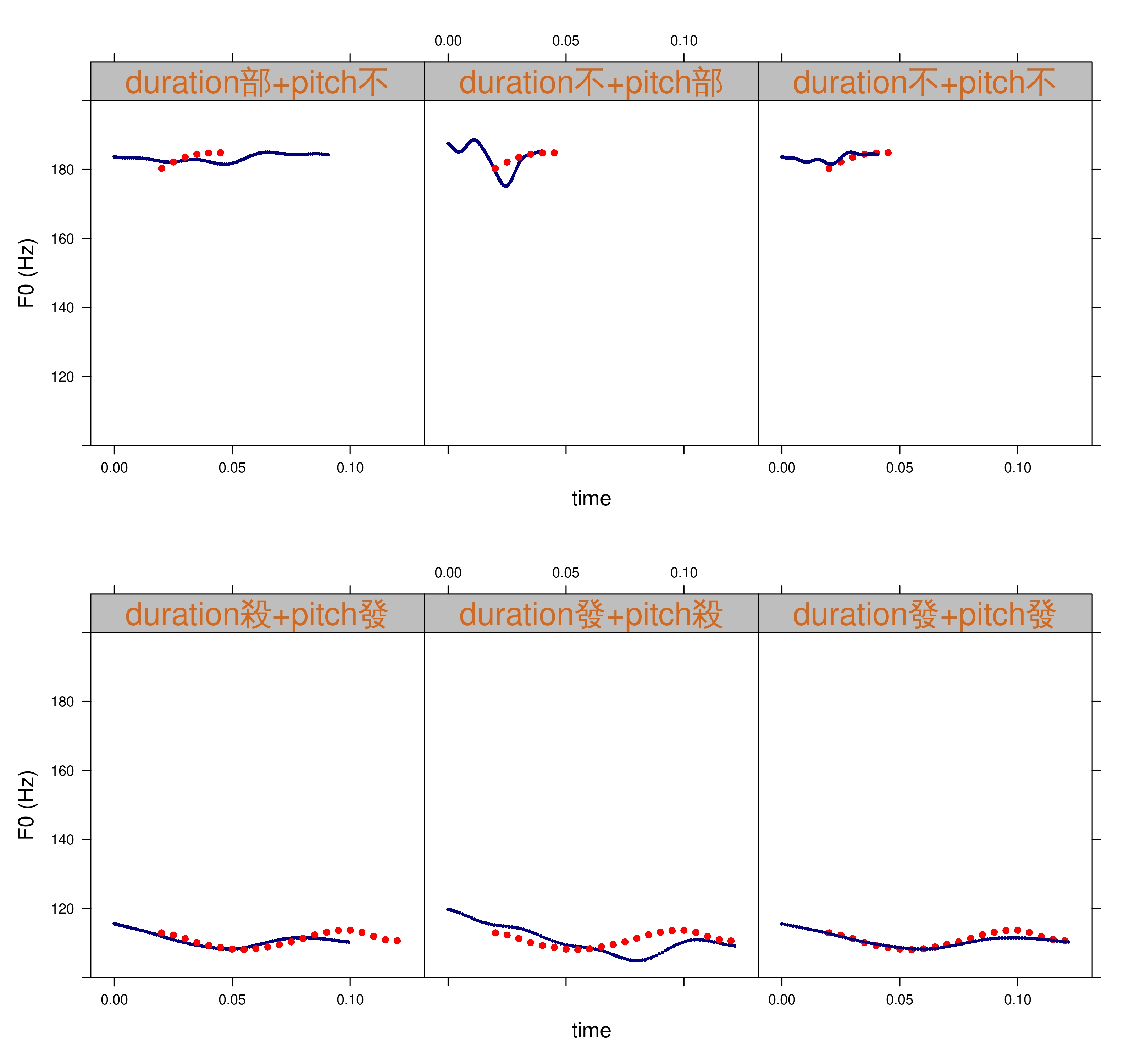}
    \caption{Examples of vowels' pitch contours in real time, predicted for the centroids of 不 and 發. Right panels: both duration and shape estimates are for 不 and 發. Left panels: duration from the homophone 部 (upper) and phonological neighbour 殺, shape from 不 and 發. Center panels: duration from 不 and 發, but shape from 部 and 殺.}
    \label{fig: bu_fa}
\end{figure}

Figure~\ref{fig: bu_fa} illustrates how predicted shape and predicted duration affect f0 contours in real time. The upper panel concerns the negation 不 (bu4, mean duration 77 ms), the lower panel concerns the verb 發 (fa1, mean duration 199 ms).  The predicted word-specific vowel duration is 40 ms for 不 and 122 ms for 發.  As the CEs are unable to capture speaker-specific differences in pitch height, we centered predicted contours around the mean of the observed contours. 
In Figure~\ref{fig: bu_fa}, the solid blue lines represent predicted pitch contours in real time.  The red dots represent the observed f0 values. In the right panels,  both pitch and duration are predicted from 不 (upper right) and 發 (lower right). In the left panels, the pitch contour is that of 不 and 發, but the duration is estimated from similar sounding words, 部 (bu4, homophonous with 不) and 殺 (sha1, same tone and vowel as 發, but a different sibilant as onset).  The center panels retain the correct durations, but take the pitch contours from 部 and 殺. These examples illustrate that the best predictions require both shape and duration to be estimated from exactly the right word. Estimates from a heterographic homophone or from a close phonological neighbour lead to marked divergences that are remarkable given that they share the same canonical tones and have the same or highly similar segments.

\section{Discussion}

We have shown that contextualized embeddings predict both spoken word duration and time-normalized f0 contours with above-chance accuracies at the type level. For spoken word duration, we report the novel finding that prediction accuracy is above chance also at the token level. 
Furthermore, combining predicted shape and duration leads to predicted f0 contours for types in real time that are 
more accurate than the contours generated from permutation baselines. 

Embeddings are, by nature, highly sensitive to context. This, of course, also holds for many aspects of prosody. The physical constraints on articulation enforce contextual smoothing that may well be, perhaps unavoidably, aligned with the in-depth co-occurrence patterns at the word level that are encoded into embeddings. This raises the question of to what extent the information encoded in contextualized embeddings reflects prosodic factors rather than word meaning in context.  We therefore considered to what extent speech rate, speaker identity, pauses in the sentence, and word type can be predicted from the contextualized embeddings. 

We first used Linear Discriminant Analysis (LDA) \cite{Fisher1936,Rao1948} to predict word type from the contextualized embeddings, to establish a semantic baseline. With leave-one-out cross-validation, accuracy (for 102 different types) was at 0.96.  This shows that there is a strong semantic component in the contextualized embeddings.   

LDA accuracy with leave-one-out cross-validation, given the task to predict speaker ($N=55$) from embeddings, was 0.637. Given that individual speakers typically discuss different topics, and given that even authors' use of function words is known to reveal their individual authorial hands, as first documented by \cite{Burrows:92}, the fact that prediction accuracy is far above a majority baseline (0.039) is unsurprising.  Importantly,  prediction accuracy for 55 speakers lags far behind prediction accuracy for 102 word types.  
 
Speech rate, as expected, is correlated with duration ($r=0.39$).  We regressed speech rate on the contextualized embeddings, and observed a correlation of predicted and observed speech rates, $r = 0.305$, that is lower than the correlation between observed and predicted spoken word duration ($r = 0.366$).  In other words, word meaning, the strongest component in the embeddings, is more closely linked to word duration than to speech rate. We note here that in GAMs predicting f0 contours in normalized time, word duration is a superior covariate compared to speech rate  \cite{Chuang:Bell:Tseng:Baayen:2024,jin2025}. 

The accuracy of an LDA given the task of predicting a pause following the word in the utterance was 0.68, which compares unfavourably to the majority baseline, 0.72.  Even though word durations tend to be longer in utterance-final position, this is not captured by the embeddings. Likewise, LDA accuracy for predicting the presence of a preceding pause, 0.83, was lower than the majority baseline (0.86). 

As part of speech is known to predict spoken word duration \cite{lohmann2018cut}, we also used LDA to predict part of speech (12 categories) from the embeddings. We used the \textbf{jiebaR} package for \texttt{R} \cite{jieba} to assign a word category to word types.  This is a rough approximation only, as Mandarin monosyllabic words tend to be used flexibly across the different parts of speech. LDA accuracy was at 0.16, lower than the majority baseline (0.20).  We leave the investigation of grammatical category at the token level for further research, and restrict ourselves to noting that parts of speech have a semantic component (e.g., nouns typically refer to entities, and verbs to actions or states) that at the token level will be confounded with the meaning of the token in context.

In this study, we modelled the effect of meaning on f0 contour shape and duration separately. It is possible that shape and duration are served by different cognitive processes. However, in general, from a purely computational perspective, two-step processing pipelines are disadvantaged compared to end-to-end pipelines.  The present approach aimed for conceptual and theoretical transparency, but the cognitive processes underlying duration and shape estimation may interact, in which case more complex models are called for.  
Nevertheless, the present results are consistent with and extending a growing body of research \cite{gahl_time_2024,plag2015homophony,lohmann2018cut,Chuang:Bell:Tseng:Baayen:2024,lu2025realization,Drager:2011,Hawkins:2003,Port:Leary:2005,de2025cracking} indicating that semantic and phonetic detail are entangled to a much greater degree than has often been assumed.

\clearpage
\newpage
\bibliographystyle{IEEEtran}
\bibliography{mybib}
\end{CJK}
\end{document}